%% file: main.tex
\documentclass{article}
\usepackage[table,xcdraw]{xcolor}
\definecolor{customblue}{RGB}{0, 102, 204}

\usepackage[final]{corl_2025} 
\usepackage{amsmath}
\usepackage{amssymb}
\usepackage{algorithm}
\usepackage{algpseudocode}
\usepackage{multirow}
\usepackage{rotating} 
\usepackage{booktabs}  
\usepackage{float}  
\usepackage{enumitem}
\usepackage[hypcap=false]{caption}
\usepackage{subcaption}

\hypersetup{
    colorlinks=true,        
    linkcolor=customblue,   
    filecolor=customblue,  
    urlcolor=customblue,   
    citecolor=customblue   
}

\title{{\huge RoboBERT: }\\
\textmd{An End-to-end Multimodal Robotic Manipulation Model}}

\author{
    Sicheng Wang$^{1,*}$ \quad 
    Sheng Liu$^{2,*}$ \quad
    Weiheng Wang$^{2,*}$ \\
    Jianhua Shan$^{3}$ \quad
    Bin Fang$^{4,\dagger}$ \\
    $^{1}$Casbot Robotic Corporation \qquad
    $^{2}$Karlsruhe Institute of Technology \qquad\\ 
    $^{3}$Anhui University of Technology \qquad
    $^{4}$Beijing Universuty of Post and Telecommunicate\\
    {\tt \small sw5425@nyu.edu, sheng.liu@student.kit.edu, utkps@student.kit.edu}
}

\begin{document}

\maketitle

\vspace{-6mm}

\input{0_abstract}    
\input{1_intro}
\input{2_related_work}
\input{3_problem}
\input{4_method}

\input{5_experiments}
\input{6_conclusion_and_limitation}






\clearpage
\bibliography{main}

\end{document}

%% file: 0_abstract.tex
\begin{abstract} 
Embodied intelligence seamlessly integrates vision, language, and action.~However, most multimodal robotic models rely on massive fine-tuning, incurring high time and hardware costs.~To address this, we introduce \textbf{RoboBERT}, an end-to-end multimodal manipulation model built around a novel two-stage training paradigm.~In the first stage, we freeze most of the vision encoder and train with a single “standard” instruction phrasing, allowing the model to focus on stable policy learning via a CNN-based diffusion policy.~In the second stage, we unfreeze all modules and inject diverse natural language variants, rapidly aligning varied instructions to the already-learned policy without destabilizing performance.~We further employ systematic data augmentations to enhance robustness against visual perturbations.~Without relying on auxiliary datasets, RoboBERT achieves new state-of-the-art (SOTA) mean episode lengths of 4.52 on the CALVIN ABCD→D benchmark and 3.79 on the ABC→D benchmark using only language-labeled expert demonstrations and a comparatively lightweight architecture.~Real-robot trials on a 6-DOF manipulator confirm higher success rates than comparable methods trained on identical data.~These results demonstrate that our data-augmentation-enhanced two-stage training paradigm delivers efficient, scalable, and broadly applicable performance for multimodal robotic systems.~More details and code are available at 
\href{https://anonymeskonto.github.io/Web/}
{\textbf{\textcolor{customblue}{anonymouskonto.github.io}}}.
\end{abstract}

%% file: 1_intro.tex
\section{Introduction}
\label{sec:intro}
In robotic manipulation, multimodal language models (MLLMs) are advancing rapidly, driven by the generalization capabilities of foundation models pretrained on internet-scale data. These models exhibit superior capacity to generalize across diverse and unseen scenarios, thereby enabling robots to execute complex tasks with remarkable adaptability \cite{firoozi2023foundationmodelsroboticsapplications}.
As a specialized class of MLLMs, Vision-Language-Action (VLA) models combine the perceptual strength of the visual modality, the semantic representation power of the language modality, and the motor execution capabilities of the action generation module, thereby forming a cohesive end-to-end decision-making framework \cite{A5,A7,yue2024deervladynamicinferencemultimodal,cheang2024gr2generativevideolanguageactionmodel,ding2025quarvlavisionlanguageactionmodelquadruped,black2024pi0visionlanguageactionflowmodel,zhen20243dvla3dvisionlanguageactiongenerative,li2024cogactfoundationalvisionlanguageactionmodel,intelligence2025pi05visionlanguageactionmodelopenworld,zheng2024tracevlavisualtraceprompting,brohan2023rt2visionlanguageactionmodelstransfer,embodimentcollaboration2024openxembodimentroboticlearning,kim2024openvlaopensourcevisionlanguageactionmodel}.


Despite these advances, LLM-style pre-training or fine-tuning is not practical for most robotic manipulation tasks due to the scarcity of large-scale high-quality action modality datasets and the heterogeneity of robot platforms \cite{geng2025roboverse}. Even if such data is available, the associated computational and time costs will be prohibitive. Furthermore, current robot-based models are highly unreliable and easily confused by different language instructions \cite{parekh2024investigatingroleinstructionvariety}. Therefore, it is crucial to maximize policy performance through careful network design and rational use of limited robot data.


To address these challenges, we propose RoboBERT, a multimodal architecture that integrates vision, language, and action through a diffusion-based policy network. To achieve robust and data-efficient learning, RoboBERT adopts a two-stage training strategy that separates stable policy formation from linguistic generalization, which will be detailed later. Beyond architecture design, we emphasize the effectiveness of systematic data augmentation in improving robustness to visual perturbations. Without relying on much robot data to fine-tune, RoboBERT achieves state-of-the-art performance on the CALVIN benchmark and demonstrates superior real-robot success rates compared to prior models.

The main contributions of our work are as follows:
\begin{enumerate}
    \item We effectively combine multiple cutting-edge technologies into a cohesive system.
    \item We propose a novel two-stage training paradigm that ensures stable and efficient policy learning while adapting to diverse natural language instructions.
    \item We develop a systematic data augmentation strategy that enhances visual robustness and generalization from limited language-labeled expert demonstrations.
    \item We conduct extensive validation of RoboBERT, achieving excellent results on both the CALVIN benchmark and real-robot experiments.
\end{enumerate}

%% file: 2_related_work.tex
\section{Related Work}
\label{sec:related_work}

End-to-end robotic control refers to the process where explicit pose estimation, grasping planning, and action planning are not required, with actions being predicted directly through learning from robot observations  \cite{A1}.
Vision-Language-Action (VLA) models, as a subclass of end-to-end robotic models, leverage the weights of a Visual-Language-Model (VLM) pre-trained on large-scale internet datasets. This approach enhances VLA’s ability to execute various language commands and allows it to generalize to new objects and scene contexts.
For instance, RT-2 \cite{brohan2023rt2visionlanguageactionmodelstransfer} fine-tunes a VLM on robot trajectory data and Internet-scale vision-language data. RT2-X \cite{embodimentcollaboration2024openxembodimentroboticlearning} trains a 55B-parameter VLA policy on the Open X-Embodiment dataset, while OpenVLA \cite{kim2024openvlaopensourcevisionlanguageactionmodel} combines a VLM backbone with more specialized robotics-related datasets. In contrast, our work proposes a two-stage trained VLA model that eliminates the need for large-scale robotics data for fine-tuning.

%% file: 3_problem.tex
\section{Problem Formulation}
\label{sec:pro}
This work implements an end-to-end, language-conditioned robotic manipulation model, where the model $\mathbf{M}$ should be able to accept natural language inputs $\mathbf{L} \in \mathbb{R}^l$ and raw observation sequences $\mathbf{V} \in \mathbb{R}^{t \times c \times h \times w}$, integrating these two different modality inputs (where \textit{t} denotes the number of time steps, \textit{c} the number of channels, and \textit{h} and \textit{w} the height and width of each image frame, respectively). The model adjusts its internal parameters \( \theta \) through a designated training method to make its output action \( \hat{\mathbf{A}} \in \mathbb{R}^a \) closely approximate the expected action \( \mathbf{A} \in \mathbb{R}^a \). This can be expressed in the formula as:
\[
\theta = \arg\min_{\theta} \left( M_{\theta}(\mathbf{L}, \mathbf{V}) - \mathbf{A} \right) = \arg\min_{\theta} \left( \hat{\mathbf{A}} - \mathbf{A} \right)
\]
To address such problems, we divide the process into three steps: the first step is feature extraction, where raw inputs are encoded, denoted as $M_{{ext-}\theta}$; the second step is the fusion of observation encodings, denoted as $M_{{fusion-}\theta}$; and the third step is policy generation, denoted as $M_{{policy-}\theta}$, which provides the probability distribution of corresponding actions by analyzing the fused results from the second step. Therefore, the model can be represented as:
\[
M_{\theta}(\mathbf{L}, \mathbf{V}) = 
M_{{policy-}\theta}\left(M_{{fusion-}\theta}\left(M_{{ext-L}\theta}(\mathbf{L}), M_{{ext-V}\theta}(\mathbf{V})\right)\right)
\]
Each module performs a specific function, and is interconnected, with their performance mutually influencing each other.

%% file: 4_method.tex
\section{Method}
\label{sec:method}

In this section, we introduce RoboBERT, a model that utilizes the BERT encoder \cite{devlin2019bertpretrainingdeepbidirectional} for language processing. We also describe the training method and data augmentation techniques tailored to optimize the model's performance.

\subsection{Model Structure}
\begin{figure*}[t] 
\centering 
\includegraphics[scale=0.51]{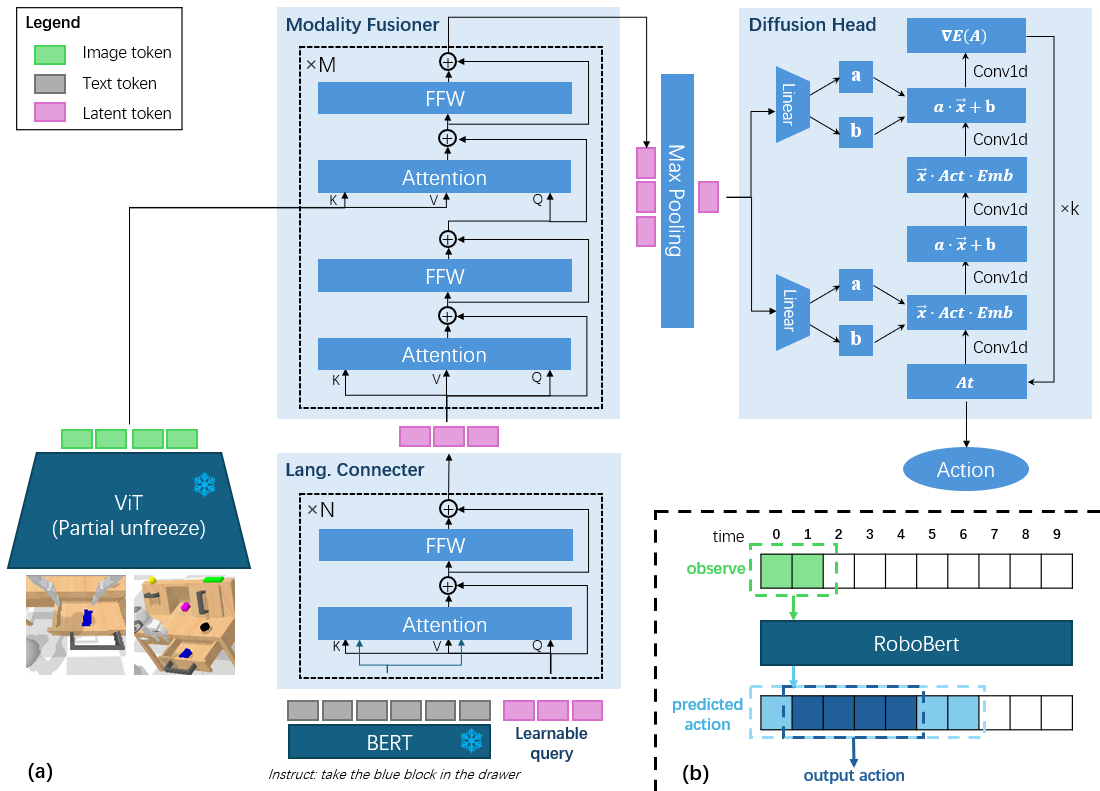}
\caption{(a) The RoboBERT architecture consists of language connectors, a modality fusion layer, and a diffusion head, responsible for sentence understanding, modality integration, and action generation, respectively. The last layer of the ViT is unfrozen during training to adapt to the task. (b) The policy workflow begins by taking observations from the last 1-2 frames, predicting actions over multiple frames, and then outputting actions for the near future. Afterward, new observations are taken, and the cycle repeats.} 
\label{1}
\end{figure*}
This section will introduce the structure of RoboBERT, which consists of three components: the feature extractor $M_{{ext-}\theta}$, the modality fusion $M_{{fusion-}\theta}$, and the action head $M_{{policy-}\theta}$. The internal implementation is shown in Fig.1, with detailed explanations provided in the subsequent sections.

\noindent
\textbf{Feature Extractor $M_{{ext-}\theta}$.}
We opted for a relatively lightweight language model BERT (110M) , often used for synonymous sentence inference. It converts the instruction input from flexible strings into a matrix of tokens. To better adapt BERT's output to subsequent stages, we add a "Language Connector", a perceiver resampler-based fine-tuning head, to further process the linguistic information and extract task-related features. For image inputs, we employ the CLIP model's Vision Transformer (ViT) \cite{A37}, trained with a text-image contrastive method that aligns language with image observations. The pre-trained CLIP model (87M) provides a strong initialization for RoboBERT training.

\noindent
\textbf{Modality Fusion $M_{{fusion-}\theta}$.}
After preprocessing by the feature extractor, the model converts various modality inputs into tokens and integrates them using a transformer decoder without a causal mask. This approach, also used in OpenFlamingo \cite{awadalla2023openflamingoopensourceframeworktraining}, treats language as the query and observations as keys and values. Fusion is achieved through multiple layers of cross-attention and self-attention. The result is a set of latent semantic tokens, which are compressed via max-pooling to represent the multimodal observation.

\noindent
\textbf{Action Head $M_{{policy-}\theta}$.}
For the action head, we use a CNN-based diffusion model \cite{A29}. This model takes a noise vector and latent representations from the modality fusion module \( \mathbf{M}_{{fusion-}\mathbf{\theta}} \) as inputs. These serve to generate seeds and denoising conditions, respectively. The action head iteratively transforms the noise seed into action vectors for a future interval. The most recent predictions are selected as the final output, as shown in Figure 1(b).

\subsection{Training Method}
\noindent
\textbf{Training Target.}
Our model is an end-to-end system that does not rely on target, pose detection, or path planning. All task-related information is contained in the original images and extracted solely by the model. We focus on the model’s inputs and outputs, using Behavioral Cloning (BC) to predict actions from observations, ensuring the agent mimics expert demonstrations under identical conditions. The weight update formula is given by:
\[
\Delta \theta = \alpha \frac{\partial L}{\partial (\mathbf{L}, \mathbf{V})} = \alpha \frac{\partial (\hat{\mathbf{A}} - \mathbf{A})^2}{\partial (\mathbf{L}, \mathbf{V})} = \alpha \frac{\partial (M_{\theta}(\mathbf{L}, \mathbf{V}) - \mathbf{A})^2}{\partial (\mathbf{L}, \mathbf{V})}
\]

However, the action head predicts noise based on polluted expert actions rather than directly predicting the action. \(M_{policy-\theta} \) predicts Gaussian noise with variance \( \epsilon^k \) from the denoising iteration \(k\) conditioned on \( M_{fusion-\theta}\), ultimately recovering the unpolluted action \(\mathbf{A}^0\). The formula will be transformed into,
\[
\Delta \theta = \alpha \frac{\partial (M_{policy-\theta}(\mathbf{A}^0 + \epsilon^k, k | M_{fusion-\theta}(\cdot)) - \epsilon^k)^2}{\partial (\mathbf{L}, \mathbf{V})}
\]

\noindent
\textbf{Two-stage Training.}
\begin{figure}[t] 
\centering 
\includegraphics[scale=0.25]{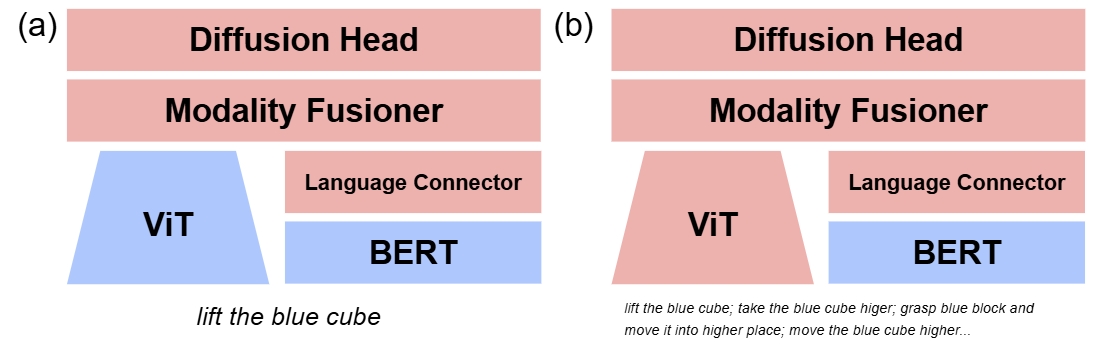}
\caption{The illustration of two-stage training. Red and blue blocks represent the activate and freezed modules (a) The first stage training, predicting the corresponding output according to the stable and simple linguistic labels. (b) The second stage training, unfreeze all the parts and train on the natural languages}
\label{1}
\end{figure}
To address the challenge of training with varied natural language inputs, we divide training into two stages. In the first stage, we use a single consistent "standard language" to stabilize label values, allowing the model to focus on learning policy. The trainable parameters and training language for this step are:
\[
\theta \in \{M_{{ext-L}\theta}, M_{{fusion-}\theta}, M_{{policy-}\theta}\}
\]
\[
\mathbf{L} \in \{\text{standard language}\}
\]

To accelerate training and preserve the pretrained weights of the CLIP encoder during the first stage, the vision encoder—typically resource-intensive—is frozen, with the exception of the final layer.

After completing the first stage, the model learns to align fixed instructions with their corresponding actions. In the second stage, we inject natural, diverse language labels to fine-tune the model, ensuring fast and accurate alignment with the previously learned policy. The second-stage loss is minimal, reducing the risk of damaging the pretrained model, and all parameters, including the vision encoder, are unfrozen for further fine-tuning.
The trainable parameters and training language for this phase are:
\[
\theta \in \{M_{{ext-L}\theta}, M_{{ext-V}\theta},M_{{fusion-}\theta}, M_{{policy-}\theta}\}
\]
\[
\mathbf{L} \in \{\text{standard language},\text{natural language}\}
\]
This two-stage approach first aligns the standard language with the policy and then aligns natural language with the standard language, ensuring seamless integration of both. The process is shown in Figure 2 (a)(b).

\subsection{Data Augmentation}
\begin{figure}[t] 
\centering 
\includegraphics[scale=0.28]{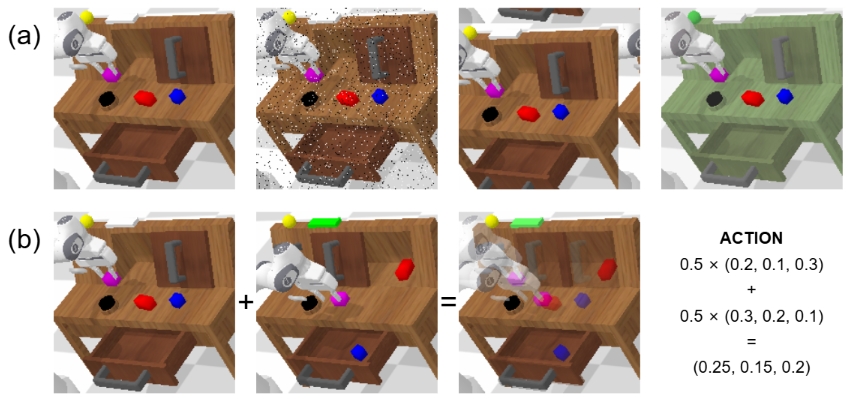}
\caption{The illustration of data augmentations. (a) From left to right, it shows the original, polluted by salt-and-pepper noise, translation, color jitter images. (b) It shows the new data generated by mixing two RGB observations and corresponding action vectors. }
\label{1}
\end{figure}
In addition, since our model directly processes raw image observations in an end-to-end manner, these images often contain a lot of redundant information, so even if the image quality degrades, the key information inside should still be captured by the model. Based on this, We emphasize the importance of the data augmentation and apply some augmentation techniques in this work.

\noindent
\textbf{Salt-and-pepper Noise.}
Salt-and-pepper noise randomly replaces pixels with black or white, simulating impulse interference. By corrupting individual pixels, it forces the model to learn holistic image features rather than rely on any single pixel (see Figure 3(a), second panel).

\noindent
\textbf{Affine Transformation.}
Real-world viewpoints vary—images can shift through translation, rotation, scaling, or distortion as the camera or robot moves—so we apply affine transformations during training, forcing the model to learn spatially invariant features from RGB inputs (see Figure 3(a), fourth panel).

\noindent
\textbf{Color Jitter.}
High-frequency features—edges, contours, shapes, and textures—carry more task-relevant information than low-frequency attributes like color or illumination. Inspired by YOLO \cite{A35}, we apply HSV color jitter during training to reduce the model’s reliance on color (see Figure 3(a), third panel). 


\noindent
\textbf{Robotic Mixup.}
Mixup \cite{A43} generates new training examples by linearly interpolating pairs of samples, boosting data diversity and generalization. In our robotic setting, we randomly select two demonstrations and compute  
\[
x_{\text{mix}} = \lambda x_0 + (1-\lambda)x_1,\quad
y_{\text{mix}} = \lambda y_0 + (1-\lambda)y_1,
\]  
where $x$ includes RGB frames and language tokens, $y$ are action vectors, and $\lambda\sim\mathrm{Beta}(0.4,0.4)$. Figure 3(b) illustrates the resulting transparent mix, with the same interpolation applied to both embeddings and actions.

%% file: 5_experiments.tex
\section{Experiments}
\label{sec:experiments}
We conducted extensive experiments on both simulation benchmarks and real robot setups to evaluate the performance of RoboBERT. Our experiments focus on the following key
questions:
\begin{enumerate}[label=\textbf{Q\arabic*)}, left=0pt]
  \item Can RoboBERT achieve comparable or superior performance to those SOTA models?
  \item Can the two-stage training paradigm improve the performance of the policy?
  \item How do different data augmentation methods affect model performance?
  
\end{enumerate}

\subsection{Simulation Evaluation}
\input{ABCDD.tex}
\input{ABCD.tex}
\noindent
\textbf{Environment and Dataset.}
We used the CALVIN dataset \cite{A20}, which is designed with a series of robotic tasks and evaluation procedures. 
CALVIN dataset includes a large number of expert demonstrations and we will use the \(ABCD \rightarrow D\) and \(ABC \rightarrow D\) subset. Additionally, only the expert demonstrations in the data include natural language descriptions of actions are used during training. According to the protocols of the environment, all the tests include 1000 groups consisting of 5 subtasks with designate natural language instructions.

\noindent
\textbf{Baselines.}
We compared our model against five baseline models: HULC \cite{A32}, GR-1 \cite{A34}, DeeR \cite{A40}, MoDE \cite{A41}, 3D diffuser Actor \cite{A42} and RoboFlamingo \cite{A5}. HULC utilizes hierarchical task representations combined with a VAE encoder to fuse multimodal information, which is used for underlying policy learning. GR-1 leverages a pre-trained model on large-scale video tasks, subsequently duplicating weights for fine-tuning on robotic tasks. RoboFlamingo utilizes a large multimodal language model as a modality fusion, using the fused latent vectors for imitation learning. DeeR-VLA introduces a multi-exit architecture in MLLMs and develops algorithms to set early-termination criteria based on demands like computational cost and GPU memory usage, which reduce the calculation. MoDE uses a diffusion policy that uses a mixture of experts transformer with a noise-conditioned routing strategy and expert caching mechanism to enhance performance and efficiency in imitation learning. 3D diffuser Actor uses a 3D denoising transformer to fuse information from 3D visual scenes, language instructions and proprioception to predict robot pose trajectory.

\noindent
\textbf{Training Configuration.}
For hardware during training, we utilized two RTX 3090s with 24GB each. When we used distributed training across two RTX 3090 24GB GPUs on dataset \(ABCD \rightarrow D\), completing ten training cycles in the first-stage training, each lasting about 40 minutes; completing five training cycles in the second-stage training, each lasting about 90 minutes.

\noindent
\textbf{Main Results.}
We compared RoboBERT against prior SOTA models in Tables 1 and 2 (S, G, P, C represent static camera, gripper camera, proprioception, and camera parameters; “Pretrain” flags use of large-scale or extra data; “Parameters” gives approximate trainable size in millions).

As observed in Table 1, which reports success rates on the \(ABCD \rightarrow D\) dataset, our model outperforms all baselines. Most of the previously top-performing methods—such as the GR-series, RoboFlamingo, and MoDE—depend on leftover unlabeled data from Calvin or additional open-source robotic datasets, resulting in higher storage requirements and training costs. Some approaches, like GR-1, incorporate extra modalities (e.g., proprioception) and 2.7 TB of sensory data, yet still achieve lower success rates than ours. Another key advantage of RoboBERT is its model size, which remains highly competitive with other state-of-the-art models such as MoDE, DeeR, and RoboFlamingo. The latter two are essentially adaptations of large foundational models, containing billions of parameters.

The table 2 shows that the performance when facing a unfamiliar D which does not appear in training dataset ABC, that is \(ABC \rightarrow D\). The model must use the experience learned from different environment to adapt on the novel environment. The result shows the similar ranking for our model on \(ABC \rightarrow D\). Comparing to the model utilizing the extra dataset or more observations input like proprioception and depth, our method still present very outstanding adaptivity and generalization, also exceeding these methods. Although our method does not have the better performance compared to MoDE with extra data, our successful rate is higher under the fair situation that using the same size of dataset. Overall, these results confirm \textbf{Q1}.

\begin{table}[t]
\centering
\begin{subtable}[t]{0.48\linewidth}
    \centering
    \scriptsize
    \begin{tabular}{llc}
        \toprule
        \textbf{Training Method} & \textbf{Dataset} & \textbf{Avg. Length} \\
        \midrule
        NL. directly & $ABCD \rightarrow D$ & $3.39 \pm 0.05$ \\
        Two-stage    & $ABCD \rightarrow D$ & $4.52 \pm 0.03$ \\
        NL. directly & $ABC \rightarrow D$  & $1.25 \pm 0.04$ \\
        Two-stage    & $ABC \rightarrow D$  & $3.79 \pm 0.03$ \\
        \bottomrule
    \end{tabular}
    \caption{Training Method}
\end{subtable}%
\hfill
\begin{subtable}[t]{0.48\linewidth}
    \centering
    \scriptsize
    \begin{tabular}{lcc}
        \toprule
        \textbf{Augmentation} & \textbf{Avg. Length} & \textbf{Increment} \\
        \midrule
        No Augmentation & $3.00 \pm 0.02$ & $+0.00$ \\
        Salt-and-Pepper & $3.22 \pm 0.03$ & $+0.22$ \\
        Affine Transformation & $2.75 \pm 0.03$ & $-0.25$ \\
        Color Jitter & $3.65 \pm 0.02$ & $+0.65$ \\
        Robotic Mixup & $3.23 \pm 0.04$ & $+0.23$ \\
        Combining All & $3.52 \pm 0.02$ & $+0.52$ \\
        Combining All w/o Aff. & $3.79 \pm 0.03$ & $+0.79$ \\
        \bottomrule
    \end{tabular}
    \caption{Data Augmentation}
\end{subtable}
\caption{Ablation on Training Method and Data Augmentation}
\label{tab:sidebyside}
\end{table}

\noindent
\textbf{Ablation on Training Method.}
To address \textbf{Q2}, we conducted the experiments that using natural languages directly (NL. directly) and two-stage training methods (Two-stage) on two dataset respectively.
The result is in Table 3 (left). We find that training directly on diverse natural-language commands paired with their corresponding actions substantially limits performance: it’s difficult for the model to learn the cross-modal correspondence all at once. By contrast, a first‐stage of simpler training yields a strong weight initialization that can reliably distinguish policies for straightforward instructions. As a result, the challenge of interpreting more complex expressions is greatly reduced, since the model already follows a correct optimization direction established during the initial, easier training.

\noindent
\textbf{Ablation on Data Augmentation.}
To address \textbf{Q3}, we conducted comparative experiments on the \(ABC \rightarrow D\) dataset using various augmentation techniques: salt-and-pepper noise (SNR = 0.95), random translation (amplitude = 15\%), Color Jitter (HSV = 0.4), robotic mixup (\(\alpha = 0.4\)) and their combinations. The augmentation parameters, such as SNR and amplitude, were empirically tuned based on trial results.

The result in Table 3 (right) shows that not all augmentations positively impact performance. For example, random translation, while providing diverse perspectives, reduced success rates. This phenomenon was also observed in the  \(ABCD \rightarrow D\) tests. We believe perspective changes may introduce spatial ambiguities; for instance, when a cube is positioned to the right of the gripper camera, the model may struggle to discern whether the cube was moved or is simply in the new position due to augmentation. While the gripper can help clarify this ambiguity, the model may still find it difficult to interpret.
Color jitter provided the most significant performance boost. Since expert demonstrations and test environments often differ in layout and color, color jitter helps decouple operations from color, focusing the model on shape instead—except for tasks where color is essential. This technique also improves robustness to variations in illumination, camera setups, shadows, and algorithm calibration in real-world scenarios. Both salt-and-pepper noise and robotic mixup give the similar and moderate gain towards the performance, suggesting that even simple and easy-implemented augmentation can improve the end-to-end robotic operations. 

For verifying their joint effectiveness, we applied all the augmentations mentioned above and conduct the test, labelled as Combining All, and all augmentations expelling the Affine Transformation giving the negative contribution, labelled as Combining All w/o Aff. The result shows that altogether improvement is undoubtedly higher than single ones but not simply additions of each increment. The result also proves that Affine Transformation do not promote the performance further. 

\subsection{Real Robot Evaluation}
\begin{figure}
    \centering
    \includegraphics[scale=0.3]{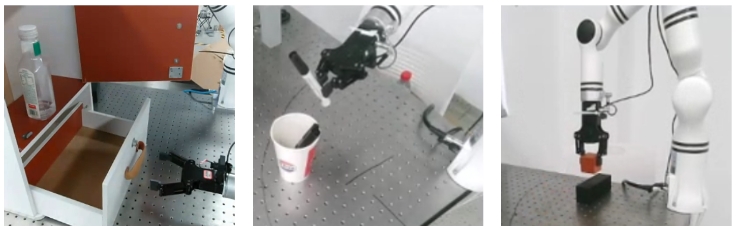}
    \caption{Some examples for real robot experiments. From left to right, sequential table tasks, moving pen to pen holder and stacking cubes.}
    \label{fig:enter-label}
\end{figure}
\input{Comparison_of_Different_Training_Methods_for_Real_Robot_Experiments.tex}
To validate RoboBERT's effectiveness in real-world environments, we conducted experiments using the REALMAN RM65B arm with 6 degrees of freedom. We designed a series of individual and instruction-following tasks to assess performance. For each task, 25-30 trajectories were collected through manual teleoperation, including RGB images from static and gripper cameras and action increments. In addition to human-generated language labels, GPT was used to generate varied expressions for the same task.

\noindent
\textbf{Individual Tasks.}
To evaluate our method’s ability to accomplish specified tasks under varying position configurations, we designed three benchmark tasks—stacking cubes (Stack Cube), returning a pen to its holder (Trans. P.), and opening a cabinet door (Open Door)—as illustrated in Figure 4. For each task, both the manipulated objects’ positions and the robot arm’s initial pose were randomized. Moreover, the test environment differed from that of the training dataset (e.g., in terms of background objects). As shown in Table 4, our model outperforms other popular language-conditioned robotic methods such as RT-1 \cite{A45} and MT-ACT \cite{A44}. 


\noindent
\textbf{Sequential Tasks.}
To assess our method’s ability to follow natural‐language instructions in long‐sequence tasks, we designed a set of sequential tasks inspired by the CALVIN benchmark. In addition to transferring objects from random start positions, these tasks include operations on articulated objects, such as opening and closing drawers. During evaluation, each action depends on the outcome of the previous step, which introduces greater uncertainty and difficulty. As shown in Table 4, our model exhibits strong potential for handling long‐horizon tasks, outperforming other state‐of‐the‐art methods.

%% file: ABCDD.tex
\begin{table*}[h]
\centering

\resizebox{1.0\textwidth}{!}{%
\begin{tabular}{@{}llcccccccc@{}}
\toprule
\textbf{Model Name} & \textbf{Observation}  &\textbf{Pertrain} & \textbf{Parameters(T)}  & \textbf{Task 1} & \textbf{Task 2} & \textbf{Task 3} & \textbf{Task 4} & \textbf{Task 5} & \textbf{Avg. Length} \\ 
\midrule

HULC&	S+G	&N	&100M &88.9\%	&73.3\%	&58.7\%	&47.5\%	&38.3\%	&$3.06\pm0.00$ \\
RoboFlamingo&	S+G	&Y &1000M	&96.4\%	 &89.6\%	&82.4\%	&74.0\%	&66.0\%	&$4.08\pm0.00$ \\
DeeR&	S+G	&Y &1000M	&98.2\%	&90.2\%	 &82.1\%	&75.9\%	&67.0\%	&$4.13\pm0.00$ \\
GR-1&	S+G+P	&Y	&130M &94.9\%	 &89.6\%	&84.4\%	&78.9\%	&73.1\%	&$4.21\pm0.00$ \\
MoDE(w. per.)&	S+G	&Y	&436M &97.1\%	&92.5\%	 &87.9\%	&83.5\%	&77.9\%	&$4.39\pm0.04$ \\
\textbf{RoboBERT(Ours)} &	S+G	&N &208M	&\textbf{98.8\%}	&\textbf{95.2\%}	 &\textbf{91.1\%}	&\textbf{86.5\%}	&\textbf{80.9\%}	&\textbf{4.52 ± 0.03} \\

\bottomrule
\end{tabular}}

\caption{CALVIN Performance Comparison on \(ABCD \rightarrow D\)}
\label{tab:robotic_models}
\end{table*}

%% file: ABCD.tex
\begin{table*}[h]
\centering

\resizebox{1.0\textwidth}{!}{%
 \begin{tabular}{@{}llcccccccc@{}}
\toprule
\textbf{Model Name} & \textbf{Observation}  &\textbf{Pertrain} & \textbf{Parameters(T)}  & \textbf{Task 1} & \textbf{Task 2} & \textbf{Task 3} & \textbf{Task 4} & \textbf{Task 5} & \textbf{Avg. Length} \\ 
\midrule

RoboFlamingo&	S+G	&Y &1000M	&82.4\%	 &61.9\%	&46.6\%	&33.1\%	&23.5\%	&$2.47\pm0.00$ \\
DeeR&	S+G	&Y &1000M	&86.2\%	&70.1\%	 &51.8\%	&41.5\%	&30.4\%	&$2.82\pm0.00$ \\
GR-1&	S+G+P	&Y	&130M &85.4\%	 &71.2\%	&59.6\%	&49.7\%	&40.1\%	&$3.06\pm0.00$ \\
3D Diffuser Actor&	S+G+P+C	&N	&200M &92.2\%	 &78.7\%	&63.9\%	&51.2\%	&41.2\%	&$3.27\pm0.04$ \\
MoDE(w/o per.)&	S+G	&N	&307M &91.5\%	&79.2\%	 &67.3\%	&55.8\%	&45.3\%	&$3.39\pm0.03$ \\

\textbf{RoboBERT(Ours)}&	S+G	&N &208M	&\textbf{95.3\%}	&\textbf{85.7\%}	 &\textbf{75.4\%}	&\textbf{66.3\%}	&\textbf{56.2\%}	&\textbf{3.79 ± 0.03} \\

\bottomrule
 \end{tabular}}
\caption{CALVIN Performance Comparison on \(ABC \rightarrow D\)}
\label{tab:robotic_models}
\end{table*}

%% file: Comparison_of_Different_Training_Methods_for_Real_Robot_Experiments.tex
\begin{table*}[t]
\centering
\resizebox{1.0\textwidth}{!}{%
\begin{tabular}{@{}ccccccccc@{}}
\hline
\multirow{2}{*}{ \textbf{Model Name}} &  \multicolumn{4}{c}{ \textbf{Sequential tasks}} & & \multicolumn{3}{c}{ \textbf{Indivdual tasks}} \\
\cline{2-5} \cline{7-9}
& Trans. D. & Trans. C. & Open D. & Close D. & & Stack Cube & Trans. P. & Open Door\\
\hline
MT-ACT & 72\% & 68\% & 73\% & 80\% &  & 72\% & 73\% & 78\%\\
RT-1 & 61\%& 56\%& 64\%& 72\% &  & 65\% & 60\% & 72\% \\
\textbf{RoboBERT(Ours)}& \textbf{86\%} & \textbf{87\%} & \textbf{80\%} & \textbf{92\%} &  & \textbf{90\%} & \textbf{80\%} & \textbf{82\%}\\
\hline

\end{tabular}}
\caption{Comparison of Different Training Methods for Real Robot Experiments}
\end{table*}

%% file: 6_conclusion_and_limitation.tex
\vspace{-0.2em}
\section{Conclusion}
\label{sec:conclusion}

This work proposes a language-conditioned multimodal robotic manipulation model, featuring a novel multiple-stage training method and data augmentation technique. The simulated and real experiments both show its competitive performance with a relatively lightweight structure and limited fine-tuning data. We hope this work will bring some beneficial inspirations to other end-to-end robotic manipulation model designs.